%% file: main.tex
\documentclass[runningheads]{llncs}

 \usepackage{eccv}

\usepackage{eccvabbrv}

\usepackage{graphicx}
\usepackage{booktabs}
\usepackage{makecell}
\usepackage{multirow}
\usepackage{pifont}

\usepackage[accsupp]{axessibility}  %

\usepackage[pagebackref,breaklinks,colorlinks,citecolor=eccvblue]{hyperref}

\usepackage{orcidlink}

\makeatletter
\renewcommand*{\@fnsymbol}[1]{\ensuremath{\ifcase#1\or *\or \dagger\or \ddagger\or
   \mathsection\or \mathparagraph\or \|\or **\or \dagger\dagger
   \or \ddagger\ddagger \else\@ctrerr\fi}}
\makeatother
\begin{document}

\title{GaussianFormer: Scene as Gaussians for Vision-Based 3D Semantic Occupancy Prediction} 

\titlerunning{GaussianFormer}

\author{Yuanhui Huang\inst{1} \and
Wenzhao Zheng\inst{1,2}\thanks{Project Leader} \and 
Yunpeng Zhang\inst{3} \and \\
Jie Zhou\inst{1} \and
Jiwen Lu\inst{1}\thanks{Corresponding Author}
}

\authorrunning{Y. Huang, W. Zheng et al.}

\institute{$^1$Tsinghua University \samelineand
$^2$University of California, Berkeley \samelineand
$^3$PhiGent Robotics\\
\url{https://wzzheng.net/GaussianFormer}\\
\texttt{huangyh22@mails.tsinghua.edu.cn; wenzhao.zheng@outlook.com;} \\
\texttt{yunpengzhang97@gmail.com; \{jzhou,lujiwen\}@tsinghua.edu.cn}
}
\newcommand{\samelineand}{\quad}

\renewcommand\twocolumn[1][]{#1}%
\maketitle
\vspace{-4mm}
\begin{center}
    \centering
    \includegraphics[width=\linewidth]{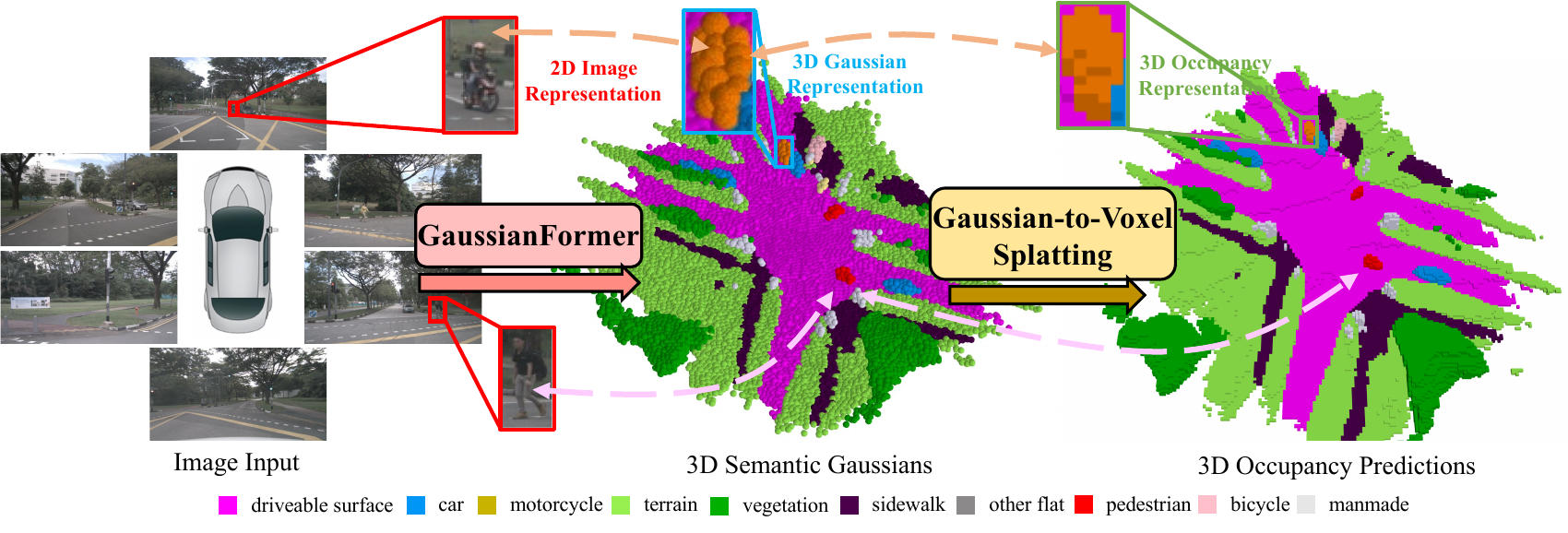}
    \vspace{-8mm}
    \captionof{figure}{
Considering the universal approximating ability of Gaussian mixture~\cite{dalal1983approximating, goodfellow2016deep}, we propose an object-centric 3D semantic Gaussian representation to describe the fine-grained structure of 3D scenes without the use of dense grids.
We propose a GaussianFormer model consisting of sparse convolution and cross-attention to efficiently transform 2D images into 3D Gaussian representations.
To generate dense 3D occupancy, we design a Gaussian-to-voxel splatting module that can be efficiently implemented with CUDA.
With comparable performance, our GaussianFormer reduces memory consumption of existing 3D occupancy prediction methods by \textbf{75.2\% - 82.2\%}.
}
\label{teaser}
\end{center}%

\input{sections/0_abstract}

\input{sections/1_introduction}

\input{sections/2_related_work}

\input{sections/3_method}
\input{sections/4_experiments}
\input{sections/5_conclusions}
\input{sections/6_appendix}

\bibliographystyle{splncs04}
\bibliography{main}
\end{document}

%% file: sections/0_abstract.tex
\begin{abstract}
3D semantic occupancy prediction aims to obtain 3D fine-grained geometry and semantics of the surrounding scene and is an important task for the robustness of vision-centric autonomous driving.
Most existing methods employ dense grids such as voxels as scene representations, which ignore the sparsity of occupancy and the diversity of object scales and thus lead to unbalanced allocation of resources.
To address this, we propose an object-centric representation to describe 3D scenes with sparse 3D semantic Gaussians where each Gaussian represents a flexible region of interest and its semantic features. 
We aggregate information from images through the attention mechanism and iteratively refine the properties of 3D Gaussians including position, covariance, and semantics.
We then propose an efficient Gaussian-to-voxel splatting method to generate 3D occupancy predictions, which only aggregates the neighboring Gaussians for a certain position.
We conduct extensive experiments on the widely adopted nuScenes and KITTI-360 datasets.
Experimental results demonstrate that GaussianFormer achieves comparable performance with state-of-the-art methods with only 17.8\% - 24.8\% of their memory consumption.
Code is available at: \url{https://github.com/huang-yh/GaussianFormer}.
\keywords{3D occupancy prediction \and 3D Gaussian splitting \and Autonomous Driving}
\end{abstract}

%% file: sections/1_introduction.tex
\section{Introduction}
\label{sec:intro}
Whether to use LiDAR for 3D perception has long been the core debate among autonomous driving companies.
While vision-centric systems share an economical advantage, their inability to capture obstacles of arbitrary shapes hinders driving safety and robustness~\cite{li2022bevformer,hu2022uniad,jiang2023vad,li2022bevdepth}. 
The emergence of 3D semantic occupancy prediction methods~\cite{cao2022monoscene,miao2023occdepth,zhang2023occformer,wei2023surroundocc,jiang2023symphonize,huang2023tri} remedies this issue by predicting the occupancy status of each voxel in the surrounding 3D space, which facilitates a variety of newly rising tasks such as end-to-end autonomous driving~\cite{occnet}, 4D occupancy forecasting~\cite{occworld}, and self-supervised 3D scene understanding~\cite{selfocc}.

Despite the promising applications, the dense output space of 3D occupancy prediction poses a great challenge in how to efficiently and effectively represent the 3D scene. 
Voxel-based methods~\cite{li2023voxformer,wei2023surroundocc} assign each voxel with a feature vector to obtain dense representations to describe the fine-grained structure of a 3D scene.
They employ coarse-to-fine upsampling~\cite{wei2023surroundocc,occnet} or voxel filtering~\cite{li2023voxformer,lu2023octreeocc} to improve efficiency considering the sparse nature of the 3D space.
As most of the voxel space is unoccupied~\cite{caesar2020nuscenes}, BEV-based methods~\cite{yu2023flashocc,li2023fb} compress the height dimension and employ the bird's eye view (BEV) as scene representations, yet they usually require post-processing such as multi-scale fusion~\cite{yu2023flashocc} to capture finer details.   
TPVFormer~\cite{huang2023tri} generalizes BEV with two additional planes and achieves a better performance-complexity trade-off with the tri-perspective view (TPV).
However, they are all grid-based methods and inevitably suffer from the redundancy of empty grids, resulting in more complexity for downstream tasks~\cite{occnet}. 
It is also more difficult to capture scene dynamics with grid-based representations since it is objects instead of grids that move in the 3D space~\cite{occworld}.

In this paper, we propose the first object-centric representation for 3D semantic occupancy prediction. 
We employ a set of 3D semantic Gaussians to sparsely describe a 3D scene.
Each Gaussian represents a flexible region of interest and consists of the mean, covariance, and its semantic category.
We propose a GaussianFormer model to effectively obtain 3D semantic Gaussians from image inputs.
We randomly initialize a set of queries to instantiate the 3D Gaussians and adopt the cross-attention mechanism to aggregate information from multi-scale image features.
We iteratively refine the properties of the 3D Gaussians for smoother optimizations.
To efficiently incorporate interactions among 3D Gaussians, we treat them as point clouds located at the Gaussian means and leverage 3D sparse convolutions to process them. 
We then decode the properties of 3D semantic Gaussians from the updated queries as the scene representation.
Motivated by the 3D Gaussian splatting method in image rendering~\cite{kerbl20233dgs}, we design an efficient Gaussian-to-voxel splatting module that aggregates neighboring Gaussians to generate the semantic occupancy for a certain 3D position.
The proposed 3D Gaussian representation uses a sparse and adaptive set of features to describe a 3D  scene but can still model the fine-grained structure due to the universal approximating ability of Gaussian mixtures~\cite{dalal1983approximating, goodfellow2016deep}.
Based on the 3D Gaussian representation, GaussianFormer further employs sparse convolution and local-aggregation-based Gaussian-to-voxel splatting to achieve efficient 3D semantic occupancy prediction, as shown in Fig.~\ref{teaser}.
We conduct extensive experiments on the nuScenes and KITTI-360 datasets for 3D semantic occupancy prediction from surrounding and monocular cameras, respectively.
GaussianFormer achieves comparable performance with existing state-of-the-art methods with only 17.8\% - 24.8\% of their memory consumption.
Our qualitative visualizations show that GaussianFormer is able to generate a both holistic and realistic perception of the scene.

%% file: sections/2_related_work.tex
\section{Related Work}

\subsection{3D Semantic Occupancy Prediction}
3D semantic occupancy prediction has garnered increasing attention in recent years due to its comprehensive description of the driving scenes, which involves predicting the occupancy and semantic states of all voxels within a certain range.
Learning an effective representation constitutes the fundamental step for this challenging task.
A straightforward approach discretizes the 3D space into regular voxels, with each voxel being assigned a feature vector~\cite{zhou2018voxelnet}. %
The capacity to represent intricate 3D structures renders voxel-based representations advantageous for 3D semantic occupancy prediction~\cite{cao2022monoscene,3DSketch,aicnet,lmscnet,yan2021JS3CNet,miao2023occdepth,zhang2023occformer,wei2023surroundocc,jiang2023symphonize}.
However, due to the sparsity of driving scenes and the high resolution of vanilla voxel representation, these approaches suffer from considerable computation and storage overhead.
To improve efficiency, several methods propose to reduce the number of voxel queries by the coarse-to-fine upsampling strategy~\cite{wang2023panoocc}, or the depth-guided voxel filtering~\cite{li2023voxformer}.
Nonetheless, the upsampling process might not adequately compensate for the information loss in the coarse stage.
And the filtering strategy ignores the occluded area and depends on the quality of depth estimation.
Although OctreeOcc~\cite{lu2023octreeocc} introduces voxel queries of multi-granularity to enhance the efficiency, it still conforms to a predefined regular partition pattern.
Our 3D Gaussian representation resembles the voxel counterpart, but can flexibly adapt to varying object scales and region complexities in a deformable way, thus achieving better resource allocation and efficiency.

Another line of work utilizes the bird's-eye-view (BEV) representation for 3D perception in autonomous driving, which can be categorized into two types according to the view transformation paradigm.
Approaches based on lift-splat-shoot (LSS) actively project image features into 3D space with depth guidance~\cite{huang2021bevdet,li2023bevstereo,li2022bevdepth,liu2023bevfusion,park2022time,philion2020lss}, while query-based methods typically use BEV queries and deformable attention to aggregate information from image features~\cite{jiang2023polarformer,li2022bevformer,yang2022bevformerv2}.
Although BEV-based perception has achieved great success in 3D object detection, it is less employed for 3D occupancy prediction due to information loss from height compression.
FB-OCC~\cite{li2023fb} uses dense BEV features from backward projection to optimize the sparse voxel features from forward projection.
FlashOcc~\cite{yu2023flashocc} applies a complex multi-scale feature fusion module on BEV features for finer details.
However, existing 3D occupancy prediction methods are based on grid representations, which inevitably suffer from the computation redundancy of empty grids.
Differently, our GaussianFormer is based on object-centric representation and can efficiently tend to flexible regions of interest.

\subsection{3D Gaussian Splatting}
The recent 3D Gaussian splatting (3D-GS)~\cite{kerbl20233dgs} uses multiple 3D Gaussians for radiance field rendering, demonstrating superior performance in rendering quality and speed. 
In contrast to prior explicit scene representations, such as meshes~\cite{qiao2023dynamic-mesh-aware,yang2022neumesh,rakotosaona2023nerfmeshing} and voxels~\cite{fridovich2022plenoxels,muller2022instant}, 3D-GS is capable of modeling intricate shapes with fewer parameters. 
Compared with implicit neural radiance field~\cite{mildenhall2021nerf,barron2021mip}, 3D-GS facilitates fast rendering through splat-based rasterization, which projects 3D Gaussians to the target 2D view and renders image patches with local 2D Gaussians.
Recent advances in 3D-GS include adaptation to dynamic scenarios~\cite{yang2023deformable-gaussian,luiten2023dynamic-3d-gaussian}, online 3D-GS generalizable to novel scenes~\cite{charatan2023pixelsplat,zou2023triplane-meets-gaussian}, and generative 3D-GS~\cite{chen2023text-to-3d-gaussian,tang2023dreamgaussian,yi2023gaussiandreamer,liang2023luciddreamer}.

Although our 3D Gaussian representation also adopts the physical form of 3D Gaussians (i.e. mean and covariance) and the multi-variant Gaussian distribution, it differs from 3D-GS in significant ways, which imposes unique challenges in 3D semantic occupancy prediction: 
1) Our 3D Gaussian representation is learned in an online manner as opposed to offline optimization in 3D-GS.
2) We generate 3D semantic occupancy predictions from the 3D Gaussian representation in contrast to rendering 2D RGB images in 3D-GS.

%% file: sections/3_method.tex
\section{Proposed Approach}

In this section, we present our method of 3D Gaussian Splatting for 3D semantic occupancy prediction.
We first introduce an object-centric 3D scene representation which adaptively describes the regions of interest with 3D semantic Gaussians (Sec.~\ref{method sub: 1}). 
We then explain how to effectively transform information from the image inputs to 3D Gaussians and elaborate on the model designs including self-encoding, image cross-attention, and property refinement (Sec.~\ref{method sub: 2}).
At last, we detail the Gaussian-to-voxel splatting module which generates dense 3D occupancy predictions based on local aggregation and can be efficiently implemented with CUDA (Sec.~\ref{method sub: 2}).

\subsection{Object-centric 3D Scene Representation}\label{method sub: 1}
\label{subsec: gaussian rep}
Vision-based 3D semantic occupancy prediction aims to predict dense occupancy states and semantics for each voxel grid with multi-view camera images as input.
Formally, given a set of multi-view images $\mathcal{I} = \{\mathbf{I}_i \in \mathbb{R}^{3\times H \times W} | i=1,...,N\}$, and corresponding intrinsics $\mathcal{K} = \{\mathbf{K}_i \in \mathbb{R}^{3\times3} | i=1,...,N\}$ and extrinsics $\mathcal{T} = \{\mathbf{T}_i \in \mathbb{R}^{4\times4} | i=1,...,N\}$, the objective is to predict 3D semantic occupancy $\mathbf{O} \in \mathcal{C}^{X\times Y\times Z}$, where $N$, $\{H, W\}$, $\mathcal{C}$ and $\{X, Y, Z\}$ denote the number of views, the image resolution, the set of semantic classes and the target volume resolution.

The autonomous driving scenes contain foreground objects of various scales (such as buses and pedestrians), and background regions of different complexities (such as road and vegetation).
Dense voxel representation~\cite{wang2023openoccupancy,cao2022monoscene,li2023voxformer} neglects this diversity and processes every 3D location with equal storage and computation resources, which often leads to intractable overhead because of unreasonable resource allocation.
Planar representations, such as BEV~\cite{huang2021bevdet,li2022bevformer} and TPV~\cite{huang2023tri}, achieve 3D perception by first encoding 3D information into 2D feature maps for efficiency and then recovering 3D structures from 2D features.
Although planar representations are resource-friendly, they could cause a loss of details.
The grid-based methods can hardly adapt to regions of interest for different scenes and thus lead to representation and computation redundancy.

\begin{figure*}[t]
\centering
\includegraphics[width=1\textwidth]{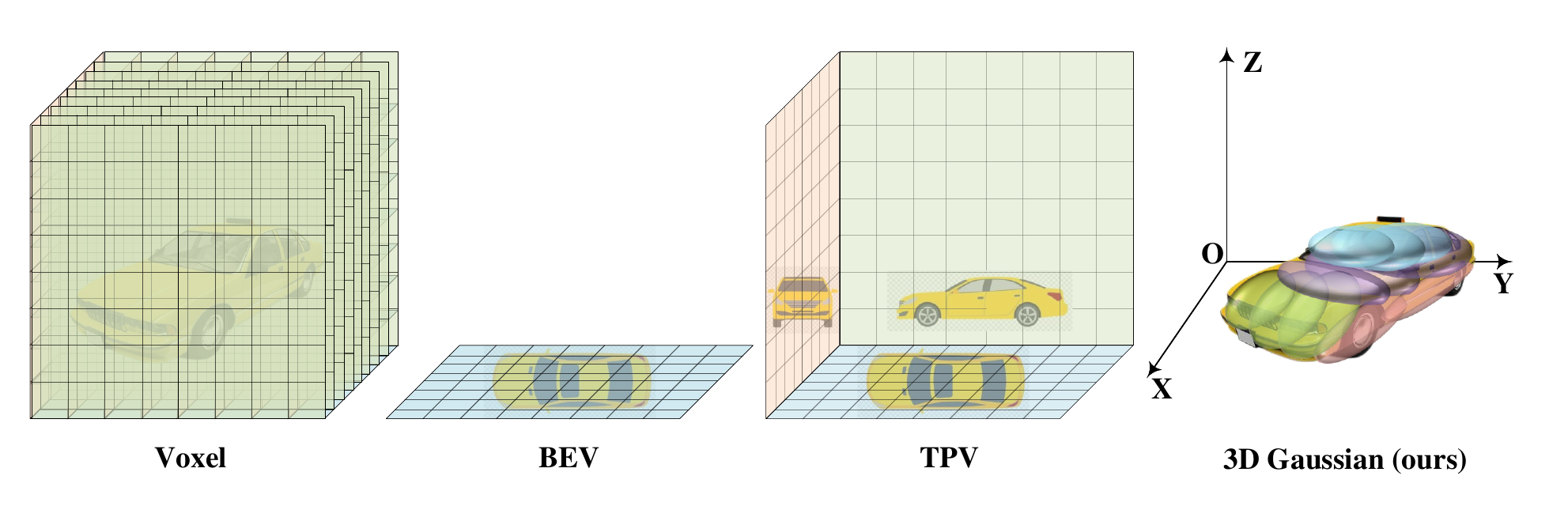}
\vspace{-9mm}
\caption{\textbf{Comparisions of the proposed 3D Gaussian representation with exiting grid-based scene representations (figures from TPVFormer~\cite{huang2023tri}).}
The voxel representation~\cite{li2023voxformer,wei2023surroundocc} assigns each voxel in the 3D space with a feature and is redundant due to the sparsity nature of the 3D space.
BEV~\cite{li2022bevformer} and TPV~\cite{huang2023tri} employ 2D planes to describe 3D space but can only alleviate the redundancy issue.
Differently, the proposed object-centric 3D Gaussian representation can adapt to flexible regions of interest yet can still describe the fine-grained structure of the 3D scene due to the strong approximating ability of mixing Gaussians~\cite{dalal1983approximating, goodfellow2016deep}.
}
\label{fig:comparison}
\vspace{-7mm}
\end{figure*}

To address this, we propose an object-centric 3D representation for 3D semantic occupancy prediction where each unit describes a region of interest instead of fixed grids, as shown in Fig.~\ref{fig:comparison}.
We represent an autonomous driving scene with a number of 3D semantic Gaussians, and each of them instantiates a semantic Gaussian distribution characterized by mean, covariance, and semantic logits.
The occupancy prediction for a 3D location can be computed by summing up the values of semantic Gaussian distributions evaluated at that location.
Specifically, we use a set of $P$ 3D Gaussians $\mathcal{G}=\{\mathbf{G}_i\in \mathbb{R}^{d} | i=1,...,P\}$ for each scene, and each 3D Gaussian is represented by a $d$-dimensional vector in the form of $(\mathbf{m}\in\mathbb{R}^{3}, \mathbf{s}\in\mathbb{R}^{3}, \mathbf{r}\in\mathbb{R}^{4}, \mathbf{c}\in\mathbb{R}^{|\mathcal{C}|})$, where $d=10+|\mathcal{C}|$, and $\mathbf{m}$, $\mathbf{s}$, $\mathbf{r}$, $\mathbf{c}$ denote the mean, scale, rotation vectors and semantic logits, respectively.
Therefore, the value of a semantic Gaussian distribution $\mathbf{g}$ evaluated at point $\mathbf{p} = (x, y, z)$ is
\begin{equation}
    \mathbf{g}(\mathbf{p}; \mathbf{m},\mathbf{s},\mathbf{r},\mathbf{c}) = {\rm{exp}}\big(-\frac{1}{2}(\mathbf{p}-\mathbf{m})^T \mathbf{\Sigma}^{-1} (\mathbf{p}-\mathbf{m})\big)\mathbf{c},
    \label{eq: gaussian dist}
\end{equation}
\begin{equation}
    \mathbf{\Sigma} = \mathbf{R}\mathbf{S}\mathbf{S}^T\mathbf{R}^T, \quad \mathbf{S} = {\rm{diag}}(\mathbf{s}), \quad \mathbf{R} = {\rm{q2r}}(\mathbf{r}),
\end{equation}
where $\mathbf{\Sigma}$, ${\rm{diag}}(\cdot)$ and ${\rm{q2r}}(\cdot)$ represent the covariance matrix, the function that constructs a diagonal matrix from a vector, and the function that transforms a quaternion into a rotation matrix, respectively.
Then the occupancy prediction result at point $\mathbf{p}$ can be formulated as the summation of the contribution of individual Gaussians on the location $\mathbf{p}$:
\begin{equation}
    \hat{o}(\mathbf{p}; \mathcal{G})=\sum_{i=1}^{P}\mathbf{g}_i(\mathbf{p}; \mathbf{m}_i,\mathbf{s}_i,\mathbf{r}_i,\mathbf{c}_i)=\sum_{i=1}^{P}{\rm{exp}}\big(-\frac{1}{2}(\mathbf{p}-\mathbf{m}_i)^T \mathbf{\Sigma}_i^{-1} (\mathbf{p}-\mathbf{m}_i)\big)\mathbf{c}_i.
    \label{eq: weighted summation}
\end{equation}

Compared with voxel representation, the mean and covariance properties allow the 3D Gaussian representation to adaptively allocate computation and storage resources according to object scales and region complexities.
Therefore, we need fewer 3D Gaussians to model a scene for better efficiency while still maintain expressiveness.
Meanwhile, the 3D Gaussian representation take 3D Gaussians as its basic unit, and thus avoids potential loss of details from dimension reduction in planar representations.
Moreover, every 3D Gaussian has explicit semantic meaning, making the transformation from the scene representation to occupancy predictions much easier than those in other representations which often involve decoding per-voxel semantics from high dimensional features.

\begin{figure*}[t]
\centering
\includegraphics[width=\textwidth]{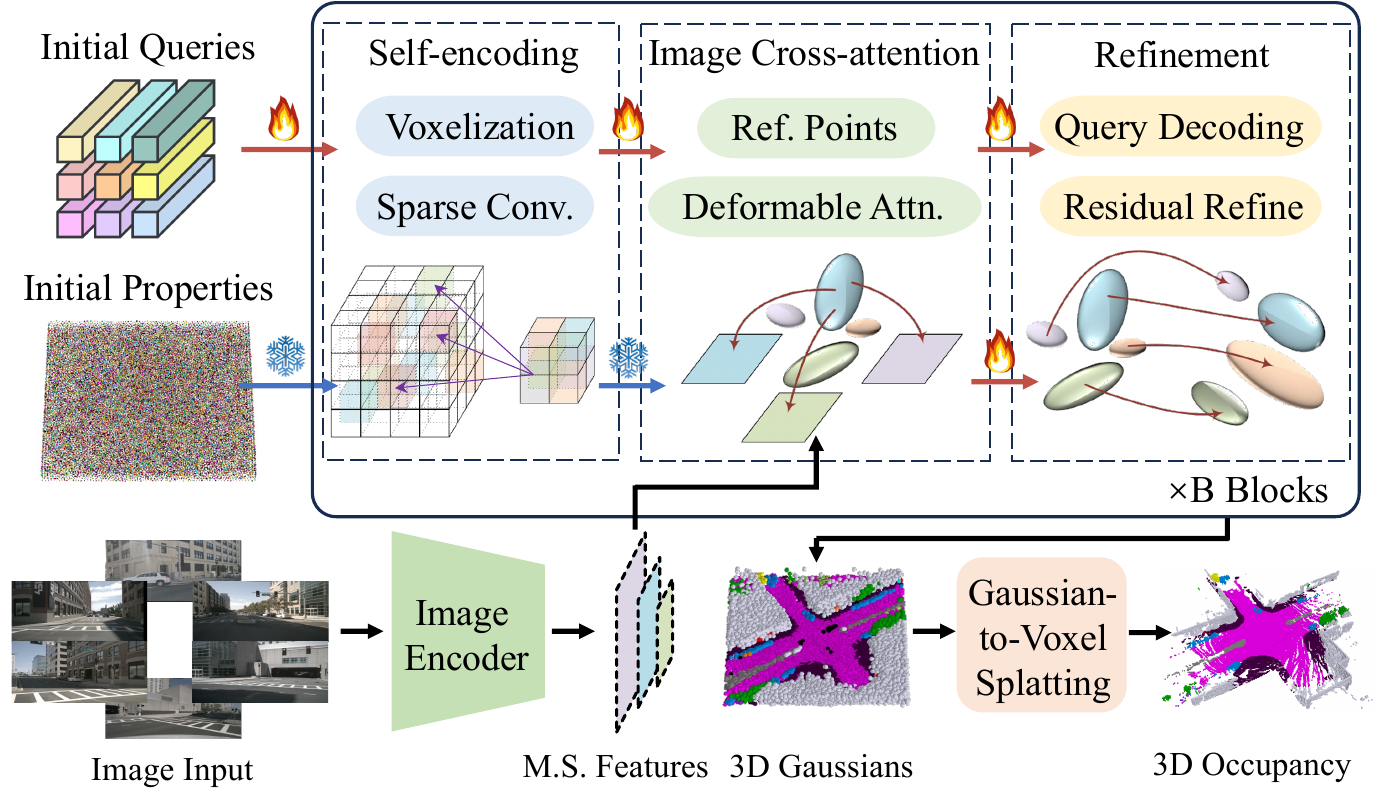}
\vspace{-7mm}
\caption{\textbf{Framework of our GaussianFormer for 3D semantic occupancy prediction.}
We first extract multi-scale (M.S.) features from image inputs using an image backbone.
We then randomly initialized a set of queries and properties (mean, covariance, and semantics) to represent 3D Gaussians and update them with interleaved self-encoding, image cross-attention, and property refinement.
Having obtained the updated 3D Gaussians, we employ an efficient Gaussian-to-voxel splatting module to generate dense 3D occupancy via local aggregation of Gaussians.
}
\label{fig:framework}
\vspace{-5mm}
\end{figure*}

\subsection{GaussianFormer: Image to Gaussians} \label{method sub: 2}
Based on the 3D semantic Gaussian representation of the scene, we further propose a GaussianFormer model to learn meaningful 3D Gaussians from multi-view images.
The overall pipeline is shown in Fig.~\ref{fig:framework}.
We first initialize the properties of 3D Gaussians and their corresponding high-dimensional queries as learnable vectors.
Then we iteratively refine the Gaussian properties within the $B$ blocks of GaussianFormer.
Each block consists of a self-encoding module to enable interactions among 3D Gaussians, an image cross-attention module to aggregate visual information, and a refinement module to rectify the properties of 3D Gaussians.

\textbf{Gaussian Properties and Queries.}
We introduce two groups of features in GaussianFormer.
The Gaussian properties $\mathcal{G}=\{\mathbf{G}_i\in \mathbb{R}^{d} | i=1,...,P\}$ are the physical attributes as discussed in Section~\ref{subsec: gaussian rep}, and they are de facto the learning target of the model.
On the other hand, the Gaussian queries $\mathcal{Q}=\{\mathbf{Q}_i\in\mathbb{R}^{m} | i=1,...,P\}$ are the high-dimensional feature vectors that implicitly encode 3D information in the self-encoding and image cross-attention modules, and provide guidance for rectification in the refinement module.
We initialize the Gaussian properties as learnable vectors denoted by Initial Properties in Fig.~\ref{fig:framework}.

\textbf{Self-encoding Module.}
Methods with voxel or planar representations usually implement self-encoding modules with deformable attention for efficiency considerations, which is not well supported for unstructured 3D Gaussian representation.
Instead, we leverage 3D sparse convolution to allow interactions among 3D Gaussians, sharing the same linear computational complexity as deformable attention.
Specifically, we treat each Gaussian as a point located at its mean $\mathbf{m}$, voxelize the generated point cloud (denoted by Voxelization in Fig.~\ref{fig:framework}), and apply sparse convolution on the voxel grid.
Since the number of 3D Gaussians $P$ is much fewer than $X\times Y\times Z$, sparse convolution could effectively take advantage of the sparsity of Gaussians.

\textbf{Image Cross-attention Module.}
The image cross-attention module (ICA) is designed to extract visual information from images for our vision-based approach.
To elaborate, for a 3D Gaussian $\mathbf{G}$, we first generate a set of 3D reference points $\mathcal{R}=\{\mathbf{m}+\Delta\mathbf{m}_i | i=1,...,R\}$ by permuting the mean $\mathbf{m}$ with offsets $\Delta\mathbf{m}$.
We calculate the offsets according to the covariance of the Gaussian to reflect the shape of its distribution.
Then we project the 3D reference points onto image feature maps with extrinsics $\mathcal{T}$ and intrinsics $\mathcal{K}$.
Finally, we update the Gaussian query $\mathbf{Q}$ with the weighted sum of retrieved image features:
\begin{equation}
    {\rm{ICA}}(\mathcal{R},\mathbf{Q},\mathbf{F}; \mathcal{T},\mathcal{K}) = \frac{1}{N}\sum_{n=1}^{N}\sum_{i=1}^{R}{\rm DA}(\mathbf{Q}, {\rm \pi}(\mathcal{R}; \mathcal{T},\mathcal{K}), \mathbf{F}_n),
\end{equation}
where $\mathbf{F}$, ${\rm DA}(\cdot)$, ${\rm \pi}(\cdot)$ denote the image feature maps, the deformable attention function and the transformation from world to pixel coordinates, respectively.

\textbf{Refinement Module.}
We use the refinement module to rectify the Gaussian properties with guidance from corresponding Gaussian queries which have aggregated sufficient 3D information in the prior self-encoding and image cross-attention modules.
Specifically, we take inspiration from DETR~\cite{zhu2020detr} in object detection.
For a 3D Gaussian $\mathbf{G}=(\mathbf{m},\mathbf{s},\mathbf{r},\mathbf{c})$, we first decode the intermediate properties $\mathbf{\hat{G}}=(\mathbf{\hat{m}},\mathbf{\hat{s}},\mathbf{\hat{r}},\mathbf{\hat{c}})$ from the Gaussian query $\mathbf{Q}$ with a multi-layer perceptron (MLP).
When refining the old properties with the intermediate ones, we treat the intermediate mean $\mathbf{\hat{m}}$ as a residual and add it with the old mean $\mathbf{m}$, while we directly substitute the other intermediate properties $(\mathbf{\hat{s}},\mathbf{\hat{r}},\mathbf{\hat{c}})$ for the corresponding old properties:
\begin{equation}
    \mathbf{\hat{G}} = (\mathbf{\hat{m}}, \mathbf{\hat{s}}, \mathbf{\hat{r}}, \mathbf{\hat{c}}) = {\rm MLP}(\mathbf{Q}), \quad \mathbf{G}_{new}=(\mathbf{m}+\mathbf{\hat{m}}, \mathbf{\hat{s}}, \mathbf{\hat{r}}, \mathbf{\hat{c}}).
\end{equation}
We refine the mean of Gaussian with residual connections in order to keep their coherence throughout the $B$ blocks of GaussianFormer.
The direct replacement of the other properties is due to the concern about vanishing gradient from the sigmoid and softmax activations we apply on covariance and semantic logits.

\subsection{Gaussian-to-Voxel Splatting} \label{method sub: 3}
Due to the universal approximating ability of Gaussian mixtures~\cite{dalal1983approximating, goodfellow2016deep}, the 3D semantic Gaussians can efficiently represent the 3D scene and thus can be directly processed to perform downstream tasks like motion planning and control.
Specifically, to achieve 3D semantic occupancy prediction, we design an efficient Gaussian-to-voxel splatting module to efficiently transform 3D Gaussian representation to 3D semantic occupancy predictions using only local aggregation operation.

\begin{figure*}[t]
\centering
\includegraphics[width=0.8\textwidth]{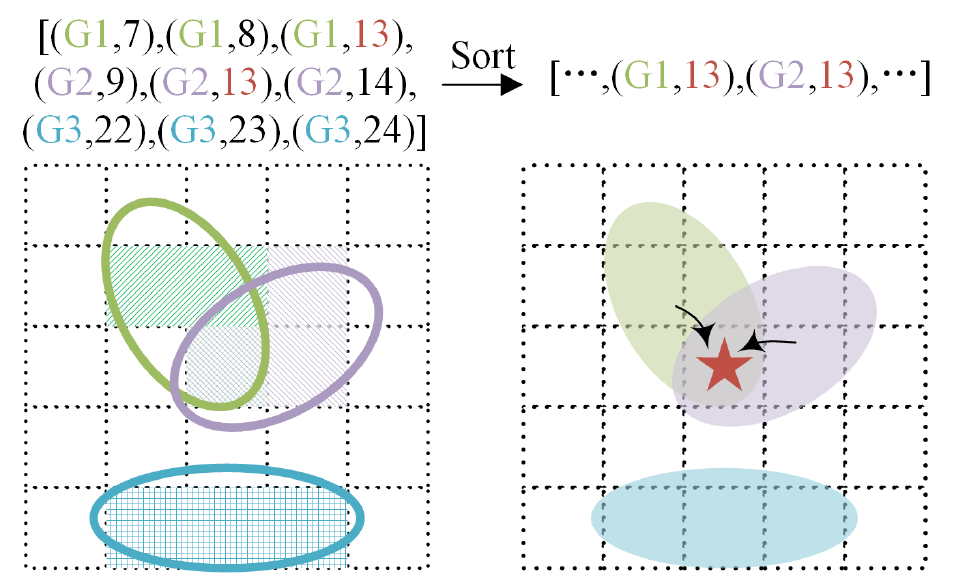}
\vspace{-2mm}
\caption{\textbf{Illustration of the Gaussian-to-voxel splatting method in 2D.}
We first voxelize the 3D Gaussians and record the affected voxels of each 3D Gaussian by appending their paired indices to a list.
Then we sort the list according to the voxel indices to identify the neighboring Gaussians of each voxel, followed by a local aggregation to generate the occupancy prediction.
}
\label{fig:gaussian-to-voxel splatting}
\vspace{-6mm}
\end{figure*}

Although \eqref{eq: weighted summation} demonstrates the main idea of the transformation as a summation over the contributions of 3D Gaussians, it is infeasible to query all Gaussians for every voxel position due to the intractable computation and storage complexity ($O(XYZ\times P)$).
Since the weight ${\rm{exp}}\big(-\frac{1}{2}(\mathbf{p}-\mathbf{m})^T \mathbf{\Sigma}^{-1} (\mathbf{p}-\mathbf{m})\big)$ in (\ref{eq: gaussian dist}) decays exponentially with respect to the square of the mahalanobis distance, it should be negligible when the distance is large enough.
Based on the locality of the semantic Gaussian distribution, we only consider 3D Gaussians within a neighborhood of each voxel position to improve efficiency.

As illustrated by Fig.~\ref{fig:gaussian-to-voxel splatting}, we first embed the 3D Gaussians into the target voxel grid of size $X\times Y\times Z$ according to their means $\mathbf{m}$.
For each 3D Gaussian, we then calculate the radius of its neighborhood according to its scale property $\mathbf{s}$.
We append both the index of the Gaussian and the index of each voxel inside the neighborhood as a tuple $(g, v)$ to a list.
Then we sort the list according to the indices of voxels to derive the indices of 3D Gaussians that each voxel should attend to:
\begin{equation}
    {\rm sort}_{vox}\big([(g, v_{g_1}),...,(g, v_{g_k})]_{g=1}^P\big) = [(g_{v_1}, v),...,(g_{v_l}, v)]_{v=1}^{XYZ},
\end{equation}
where $k$, $l$ denote the number of neighboring voxels of a certain Gaussian, and the number of Gaussians that contribute to a certain voxel, respectively.
Finally, we can approximate (\ref{eq: weighted summation}) efficiently with only neighboring Gaussians:
\begin{equation}
    \hat{o}(\mathbf{p}; \mathcal{G})=\sum_{i\in \mathcal{N}(\mathbf{p})}\mathbf{g}_i(\mathbf{p}; \mathbf{m}_i,\mathbf{s}_i,\mathbf{r}_i,\mathbf{c}_i),
    \label{eq: neighbor weighted summation}
\end{equation}
where $\mathcal{N}(\mathbf{p})$ represents the set of neighboring Gaussians of the point at $\mathbf{p}$.
Considering the dynamic neighborhood sizes of 3D Gaussians, the implementation of Gaussian-to-voxel splatting is nontrivial.
To fully exploit the parallel computation ability of GPU, we realize it with the CUDA programming language to achieve better acceleration.

The overall GaussianFormer model can be trained efficiently in an end-to-end manner. 
For training, we use the cross entropy loss $L_{ce}$ and the lovasz-softmax~\cite{berman2018lovasz} loss $L_{lov}$ following TPVFormer~\cite{huang2023tri}.
To refine the Gaussian properties in an iterative manner, we apply supervision on the output of each refinement module. 
The overall loss function is $L=\sum_{i=1}^{B}L_{ce}^{i} + L_{lov}^{i}$, where $i$ denote the $i$-th block.

%% file: sections/4_experiments.tex
\section{Experiments}
\definecolor{nbarrier}{RGB}{255, 120, 50}
\definecolor{nbicycle}{RGB}{255, 192, 203}
\definecolor{nbus}{RGB}{255, 255, 0}
\definecolor{ncar}{RGB}{0, 150, 245}
\definecolor{nconstruct}{RGB}{0, 255, 255}
\definecolor{nmotor}{RGB}{200, 180, 0}
\definecolor{npedestrian}{RGB}{255, 0, 0}
\definecolor{ntraffic}{RGB}{255, 240, 150}
\definecolor{ntrailer}{RGB}{135, 60, 0}
\definecolor{ntruck}{RGB}{160, 32, 240}
\definecolor{ndriveable}{RGB}{255, 0, 255}
\definecolor{nother}{RGB}{139, 137, 137}
\definecolor{nsidewalk}{RGB}{75, 0, 75}
\definecolor{nterrain}{RGB}{150, 240, 80}
\definecolor{nmanmade}{RGB}{213, 213, 213}
\definecolor{nvegetation}{RGB}{0, 175, 0}

\newcommand\crule[3][black]{\textcolor{#1}{\rule{#2}{#3}}}
\definecolor{nvcolor}{RGB}{119,185,0}
\definecolor{roadcolor}{RGB}{234,51,246}
\definecolor{sidewalkcolor}{RGB}{68,8,72}
\definecolor{parkingcolor}{RGB}{241,156,249}
\definecolor{othergroundcolor}{RGB}{160,32,76}
\definecolor{buildingcolor}{RGB}{246,202,69}
\definecolor{carcolor}{RGB}{111,149,238}
\definecolor{truckcolor}{RGB}{74,32,172}
\definecolor{bicyclecolor}{RGB}{136,227,242}
\definecolor{motorcyclecolor}{RGB}{37,59,146}
\definecolor{othervehiclecolor}{RGB}{96,81,242}
\definecolor{vegetationcolor}{RGB}{79, 173, 50}
\definecolor{trunkcolor}{RGB}{126, 65, 22}
\definecolor{terraincolor}{RGB}{171, 238, 105}
\definecolor{personcolor}{RGB}{234, 60, 49}
\definecolor{bicyclistcolor}{RGB}{234, 66, 195}
\definecolor{motorcyclistcolor}{RGB}{138, 42, 90}
\definecolor{fencecolor}{RGB}{238, 128, 69}
\definecolor{polecolor}{RGB}{252, 241, 161}
\definecolor{trafficsigncolor}{RGB}{233, 51, 35}
\definecolor{other-struct.color}{RGB}{255, 150, 0}
\definecolor{other-objectcolor}{RGB}{50, 255, 255}
\definecolor{lane-markingcolor}{RGB}{150, 255, 170}
\definecolor{color1}{RGB}{176, 36, 24}
\definecolor{color2}{RGB}{0, 176, 80}
\definecolor{color3}{RGB}{0, 0, 200}
\newcommand{\tbr}[1]{\textbf{\textcolor{color1}{#1}}}
\newcommand{\tbg}[1]{\textbf{\textcolor{color2}{#1}}}
\newcommand{\tbb}[1]{\textbf{\textcolor{color3}{#1}}}

In this paper, we propose a 3D semantic Gaussian representation to effectively and efficiently describe the 3D scene and devise a GaussianFormer model to perform 3D occupancy prediction.
We conducted experiments on the nuScenes~\cite{caesar2020nuscenes} dataset and the KITTI-360~\cite{Liao2022kitti360} dataset for 3D semantic occupancy prediction with surrounding and monocular cameras, respectively.

\subsection{Datasets}
\textbf{NuScenes}~\cite{caesar2020nuscenes} consists of 1000 sequences of various driving scenes collected in Boston and Singapore, which are officially split into 700/150/150 sequences for training, validation and testing, respectively. 
Each sequence lasts 20 seconds with RGB images collected by 6 surrounding cameras, and the keyframes are annotated at 2Hz. 
We leverage the dense semantic occupancy annotations from SurroundOcc~\cite{wei2023surroundocc} for supervision and evaluation.
The annotated voxel grid spans [-50m, 50m] along the X and Y axes, and [-5m, 3m] along the Z axis with a resolution of $200\times 200\times 16$.
Each voxel is labeled with one of the 18 classes (16 semantic, 1 empty and 1 unknown classes).

\textbf{KITTI-360}~\cite{Liao2022kitti360} is a large-scale dataset covering a driving distance of 73.7km corresponding to over 320k images and 100k laser scans.
We use the dense semantic occupancy annotations from SSCBench-KITTI-360~\cite{li2023sscbench} for supervision and evaluation.
It provides ground truth labels for 9 long sequences with a total of 12865 key frames, which are officially split into 7/1/1 sequences with 8487/1812/2566 key frames for training, validation and testing, respectively.
The voxel grid spans $51.2\times 51.2\times 6.4 m^3$ in front of the ego car with a resolution of $256 \times 256 \times 32$, and each voxel is labeled with one of 19 classes (18 semantic and 1 empty).
We use the RGB images from the left camera as input to our model.

\subsection{Evaluation Metrics}

Following common practice~\cite{cao2022monoscene}, we use mean Intersection-over-Union (mIoU) and Intersection-over-Union (IoU) to evaluate the performance of our model:
\begin{equation}
    {\rm mIoU} = \frac{1}{|\mathcal{C}'|}\sum_{i\in\mathcal{C}'} \frac{TP_i}{TP_i+FP_i+FN_i}, \quad {\rm IoU}=\frac{TP_{\neq c_0}}{TP_{\neq c_0}+FP_{\neq c_0}+FN_{\neq c_0}},
\end{equation}
where $\mathcal{C}'$, $c_0$, $TP$, $FP$, $FN$ denote the nonempty classes, the empty class, the number of true positive, false positive and false negative predictions, respectively. 

\subsection{Implementation Details}
We set the resolutions of input images as $900\times 1600$ for nuScenes~\cite{caesar2020nuscenes} and $376\times 1408$ for KITTI-360~\cite{Liao2022kitti360}.
We employ ResNet101-DCN~\cite{he2016resnet} initialized from FCOS3D~\cite{wang2021fcos3d} checkpoint as the image backbone for nuScenes and ResNet50~\cite{he2016resnet} pretrained with ImageNet~\cite{deng2009imagenet} for KITTI-360.
We use the feature pyramid network~\cite{lin2017fpn} (FPN) to generate multi-scale image features with downsample rates of 4, 8, 16 and 32.
We set the number of Gaussians to 144000 and 38400 for nuScenes and KITTI-360, respectively, and use 4 transformer blocks in GaussianFormer to refine the properties of Gaussians.
For optimization, we utilize the AdamW~\cite{loshchilov2017adamw} optimizer with a weight decay of 0.01.
The learning rate warms up in the first 500 iterations to a maximum value of 2e-4 and decreases according to a cosine schedule.
We train our models for 20 epochs with a batch size of 8, and employ random flip and photometric distortion augmentations.

\begin{table*}[t] %
    \caption{\textbf{3D semantic occupancy prediction results on nuScenes validation set.} While the original TPVFormer~\cite{huang2023tri} is trained with LiDAR segmentation labels, TPVFormer* is supervised by dense occupancy annotations. Our method achieves comparable performance with state-of-the-art methods.}
    \setlength{\tabcolsep}{0.003\linewidth}  
    \vspace{-3mm}  
    \renewcommand\arraystretch{1.3}
    \centering
    \resizebox{\textwidth}{!}{
    \begin{tabular}{l|c c | c c c c c c c c c c c c c c c c}
        \toprule
        Method
        &  \makecell{SC\\ IoU} & \makecell{SSC \\ mIoU}
        & \rotatebox{90}{\textcolor{nbarrier}{$\blacksquare$} barrier}
        & \rotatebox{90}{\textcolor{nbicycle}{$\blacksquare$} bicycle}
        & \rotatebox{90}{\textcolor{nbus}{$\blacksquare$} bus}
        & \rotatebox{90}{\textcolor{ncar}{$\blacksquare$} car}
        & \rotatebox{90}{\textcolor{nconstruct}{$\blacksquare$} const. veh.}
        & \rotatebox{90}{\textcolor{nmotor}{$\blacksquare$} motorcycle}
        & \rotatebox{90}{\textcolor{npedestrian}{$\blacksquare$} pedestrian}
        & \rotatebox{90}{\textcolor{ntraffic}{$\blacksquare$} traffic cone}
        & \rotatebox{90}{\textcolor{ntrailer}{$\blacksquare$} trailer}
        & \rotatebox{90}{\textcolor{ntruck}{$\blacksquare$} truck}
        & \rotatebox{90}{\textcolor{ndriveable}{$\blacksquare$} drive. suf.}
        & \rotatebox{90}{\textcolor{nother}{$\blacksquare$} other flat}
        & \rotatebox{90}{\textcolor{nsidewalk}{$\blacksquare$} sidewalk}
        & \rotatebox{90}{\textcolor{nterrain}{$\blacksquare$} terrain}
        & \rotatebox{90}{\textcolor{nmanmade}{$\blacksquare$} manmade}
        & \rotatebox{90}{\textcolor{nvegetation}{$\blacksquare$} vegetation}
        \\
        \midrule
        MonoScene~\cite{cao2022monoscene} & 23.96 & 7.31 & 4.03 &	0.35& 8.00& 8.04&	2.90& 0.28& 1.16&	0.67&	4.01& 4.35&	27.72&	5.20& 15.13&	11.29&	9.03&	14.86 \\
        
        Atlas~\cite{murez2020atlas} & 28.66 & 15.00 & 10.64&	5.68&	19.66& 24.94& 8.90&	8.84&	6.47& 3.28&	10.42&	16.21&	34.86&	15.46&	21.89&	20.95&	11.21&	20.54 \\
        
        BEVFormer~\cite{li2022bevformer} & 30.50 & 16.75 & 14.22 &	6.58 & 23.46 & 28.28& 8.66 &10.77& 6.64& 4.05& 11.20&	17.78 & 37.28 & 18.00 & 22.88 & 22.17 & \tbg{13.80} &	\tbr{22.21}\\
        
        TPVFormer~\cite{huang2023tri} & 11.51 & 11.66 & 16.14&	7.17& 22.63	& 17.13 & 8.83 & 11.39 & 10.46 & 8.23&	9.43 & 17.02 & 8.07 & 13.64 & 13.85 & 10.34 & 4.90 & 7.37\\
        
        TPVFormer*~\cite{huang2023tri}  & \tbb{30.86} & 17.10 & 15.96&	 5.31& 23.86	& 27.32 & 9.79 & 8.74 & 7.09 & 5.20& 10.97 & 19.22 & \tbg{38.87} & \tbb{21.25} & \tbb{24.26} & \tbr{23.15} & 11.73 & 20.81\\

        OccFormer~\cite{zhang2023occformer} & \tbg{31.39} & \tbb{19.03} & \tbb{18.65} & \tbb{10.41} & \tbb{23.92} & \tbg{30.29} & \tbb{10.31} & \tbg{14.19} & \tbg{13.59} & \tbg{10.13} & \tbb{12.49} & \tbb{20.77} & \tbb{38.78} & 19.79 & 24.19 & 22.21 & \tbb{13.48} & \tbb{21.35}\\
        
        SurroundOcc~\cite{wei2023surroundocc} & \tbr{31.49} & \tbr{20.30}  & \tbr{20.59} & \tbr{11.68} & \tbr{28.06} & \tbr{30.86} & \tbr{10.70} & \tbr{15.14} & \tbr{14.09} & \tbr{12.06} & \tbr{14.38} & \tbr{22.26} & 37.29 & \tbr{23.70} & \tbr{24.49} & \tbb{22.77} & \tbr{14.89} & \tbg{21.86}  \\ 

        \midrule

        \textbf{Ours} & 29.83 & \tbg{19.10} & \tbg{19.52} & \tbg{11.26} & \tbg{26.11} & \tbb{29.78} & \tbg{10.47} & \tbb{13.83} & \tbb{12.58} & \tbb{8.67} & \tbg{12.74} & \tbg{21.57} & \tbr{39.63} & \tbg{23.28} & \tbg{24.46} & \tbg{22.99} & 9.59 & 19.12 \\
        
        \bottomrule
    \end{tabular}}
    \label{tab:nuscseg}
    \vspace{-3mm}
\end{table*}

\begin{table*}[t!]
    \centering
    \caption{\textbf{3D semantic occupancy prediction results on SSCBench-KITTI-360 validation set.}
    Our method achieves performance on par with state-of-the-art methods, excelling at some smaller and general categories (i.e. motorcycle, other-veh.).}
    \label{tab:kittiseg}
    \vspace{-3mm}
    \setlength{\tabcolsep}{0.003\linewidth}   
    \renewcommand\arraystretch{1.3}
    \resizebox{1\linewidth}{!}{
    \begin{tabular}{l|c|c|c| c c c c c c c c c c c c c c c c c c}
    \toprule
    Method & {\rotatebox{90}{Input}}  & \makecell{SC\\IoU} & \makecell{SSC\\mIoU}
    &\rotatebox{90}{\textcolor{carcolor}{$\blacksquare$} car}
    &\rotatebox{90}{\textcolor{bicyclecolor}{$\blacksquare$} {bicycle}}
    &\rotatebox{90}{\textcolor{motorcyclecolor}{$\blacksquare$} {motorcycle}}
    &\rotatebox{90}{\textcolor{truckcolor}{$\blacksquare$} {truck}}
    &\rotatebox{90}{\textcolor{othervehiclecolor}{$\blacksquare$} {other-veh.}}
    &\rotatebox{90}{\textcolor{personcolor}{$\blacksquare$} {person}}
    &\rotatebox{90}{\textcolor{roadcolor}{$\blacksquare$} {road}}  
    &\rotatebox{90}{\textcolor{parkingcolor}{$\blacksquare$} {parking}}
    &\rotatebox{90}{\textcolor{sidewalkcolor}{$\blacksquare$} {sidewalk}}
    &\rotatebox{90}{\textcolor{othergroundcolor}{$\blacksquare$} {other-grnd}}
    &\rotatebox{90}{\textcolor{buildingcolor}{$\blacksquare$} {building}}
    &\rotatebox{90}{\textcolor{fencecolor}{$\blacksquare$} {fence}}
    &\rotatebox{90}{\textcolor{vegetationcolor}{$\blacksquare$} {vegetation}}
    &\rotatebox{90}{\textcolor{terraincolor}{$\blacksquare$} {terrain}}
    &\rotatebox{90}{\textcolor{polecolor}{$\blacksquare$} {pole}}
    &\rotatebox{90}{\textcolor{trafficsigncolor}{$\blacksquare$} {traf.-sign}}
    &\rotatebox{90}{\textcolor{other-struct.color}{$\blacksquare$} {other-struct.}}
    &\rotatebox{90}{\textcolor{other-objectcolor}{$\blacksquare$} {other-object}}
     
    \\\midrule

    LMSCNet~\cite{lmscnet} & L & {47.53} & {13.65} & {20.91} & {0} & {0} & {0.26} & {0} & {0} & {62.95} & {13.51} & {33.51} & {0.2} & {43.67} & {0.33} & {40.01} & {26.80} & {0} & {0} & {3.63} & {0}
        
    \\ SSCNet~\cite{song2017semantic} & L & {53.58} & {16.95} & {31.95} & {0} & {0.17} & {10.29} & {0.58} & {0.07} & {65.7} & {17.33} & {41.24} & {3.22} & {44.41} & {6.77} & {43.72} & {28.87} & {0.78} & {0.75} & {8.60} & {0.67} 
    
    \\\midrule MonoScene~\cite{cao2022monoscene} & C & {37.87} & {12.31} & {19.34} & {0.43} & {0.58} & {8.02} & {2.03} & {0.86} & {48.35} & {11.38} & {28.13} & {3.22} & {32.89} & {3.53} & {26.15} & \tbb{16.75} & {6.92} & {5.67} & {4.20} & {3.09}
    
    \\ Voxformer~\cite{li2023voxformer} & C & {38.76} & {11.91} & {17.84} & \tbg{1.16} & {0.89}& {4.56} & {2.06}  & {1.63} & {47.01} & {9.67} & {27.21} & {2.89} & {31.18} & \tbb{4.97} & {28.99} & {14.69} & {6.51} & \tbb{6.92} & {3.79} & {2.43}
    
    \\ TPVFormer~\cite{huang2023tri} & C & \tbb{40.22} & \tbb{13.64} & \tbb{21.56} & \tbb{1.09} & \tbb{1.37} & {8.06} & {2.57} & {2.38} & \tbb{52.99} & \tbb{11.99} & \tbb{31.07} & \tbb{3.78} & \tbb{34.83} & {4.80} & \tbb{30.08} & \tbg{17.51} & \tbb{7.46} & {5.86} & {5.48} & {2.70} 
    
    \\ OccFormer~\cite{zhang2023occformer} & C & \tbg{40.27} & \tbg{13.81} & \tbg{22.58} & {0.66} & {0.26} & \tbb{9.89} & \tbb{3.82} & \tbb{2.77} & \tbg{54.30} & \tbg{13.44} & \tbg{31.53} & {3.55} & \tbr{36.42} & {4.80} & \tbg{31.00} & \tbr{19.51} & \tbg{7.77} & \tbg{8.51} & \tbb{6.95} & \tbb{4.60} 

    \\ Symphonies~\cite{jiang2023symphonize} & C & \tbr{44.12} & \tbr{18.58} & \tbr{30.02} & \tbr{1.85} & \tbr{5.90} & \tbr{25.07} & \tbr{12.06} & \tbr{8.20} & \tbr{54.94} & \tbr{13.83} & \tbr{32.76} & \tbr{6.93} & \tbg{35.11} & \tbr{8.58} & \tbr{38.33} & 11.52 & \tbr{14.01} & \tbr{9.57} & \tbr{14.44} & \tbr{11.28}

    \\\midrule \textbf{Ours} & C & 35.38 & {12.92} & 18.93 & {1.02} & \tbg{4.62} & \tbg{18.07} & \tbg{7.59} & \tbg{3.35} & 45.47 & 10.89 & 25.03 & \tbg{5.32} & 28.44 & \tbg{5.68} & {29.54} & 8.62 & 2.99 & 2.32 & \tbg{9.51} & \tbg{5.14}
 
\\\bottomrule
\end{tabular}
}
\vspace{-7mm}
\end{table*}

\subsection{Results and Analysis}

\textbf{Surrounding-Camera 3D semantic occupancy prediction.}
In Table~\ref{tab:nuscseg}, we present a comprehensive quantitative comparison of various methods for multi-view 3D semantic occupancy prediction on nuScenes validation set, with dense annotations from SurroundOcc~\cite{wei2023surroundocc}.
Our GaussianFormer achieves notable improvements over methods based on planar representations, such as BEVFormer~\cite{li2022bevformer} and TPVFormer~\cite{huang2023tri}.
Even compared with dense grid representations, GaussianFormer performs on par with OccFormer~\cite{zhang2023occformer} and SurroundOcc~\cite{wei2023surroundocc}.
These observations prove the valuable application of the 3D Gaussians for semantic occupancy prediction.
This is because the 3D Gaussian representation better exploits the sparse nature of the driving scenes and the diversity of object scales with flexible properties of position and covariance.

\textbf{Monocular 3D semantic occupancy prediction.}
Table~\ref{tab:kittiseg} compares the performance of GaussianFormer with other methods for monocular 3D semantic occupancy prediciton on SSCBench-KITTI-360.
Notably, GaussianFormer achieves comparable performance with state-of-the-art models, excelling at some smaller categories such as motorcycle and general categories such as other-vehicle.
This is due to 3D Gaussians can adaptively change their positions and covariance to match the boundaries of small objects in images in contrast to rigid grid projections on images which might be misleading.
Furthermore, the flexibility of 3D Gaussians also benefits the predictions for general objects (i.e. categories with other- prefix) which often have distinct shapes and appearances with normal categories.

\textbf{Efficiency comparisons with existing methods.}
We provide the efficiency comparisons of different scene representations in Table~\ref{tab:efficacy}.
Notably, GaussianFormer surpasses all existing competitors with significantly reduced memory consumption.
The memory efficiency of GaussianFormer originates from its object-centric nature which assigns explicit semantic meaning to each 3D Gaussian, and thus greatly simplifies the transformation from the scene representation to occupancy predictions, getting rid of the expensive decoding process from high dimensional features.
While being slightly slower ($\sim$ 70 ms) than methods based on planar representations~\cite{li2022bevformer,huang2023tri}, GaussianFormer achieves the lowest latency among dense grid representations.
Notably, our method is faster than OctreeOcc~\cite{lu2023octreeocc} even with more queries.

\begin{table*}[t] %
    \centering
    \caption{\textbf{Efficiency comparison of different representations} on nuScenes. 
    The latency and memory consumption for GaussianFormer are tested on one NVIDIA 4090 GPU with batch size one, while the results for other methods are reported in OctreeOcc~\cite{lu2023octreeocc} tested on one NVIDIA A100 GPU.
    Our method demonstrates significantly reduced memory usage compared to other representations.}
    \vspace{-3mm} 
    \setlength{\tabcolsep}{0.01\linewidth}    
    \resizebox{1\linewidth}{!}{
    \begin{tabular}{l|c|c|cc}
    \toprule
    Methods & Query Form & Query Resolution & Latency~$\downarrow$ & Memory~$\downarrow$ \\ \midrule
    BEVFormer~\cite{li2022bevformer} & 2D BEV & 200$\times$200 & \textbf{302} ms   &25100 M \\
    TPVFormer~\cite{huang2023tri} & 2D tri-plane & 200$\times$(200+16+16) & 341 ms & 29000 M \\
    PanoOcc~\cite{wang2023panoocc} & 3D voxel & 100$\times$100 $\times$16 & 502 ms & 35000 M \\
    FBOCC~\cite{li2023fb} & 3D voxel \& 2D BEV & 200$\times$200$\times$16 \& 200$\times$200 & 463 ms & 31000 M \\
    OctreeOcc~\cite{lu2023octreeocc} & Octree Query & 91200 & 386 ms & 26500 M \\\midrule
    GaussianFormer & 3D Gaussian & 144000 & 372 ms & \textbf{6229} M \\
    \bottomrule
    \end{tabular}}
    \vspace{-3mm}
\label{tab:efficacy}
\end{table*}

\begin{table*}[t!]
    \centering
    \caption{
    \textbf{Ablation on the components of GaussianFormer.}
    Deep Supervision represents supervising the output of each refinement module.
    Residual Refine means on which properties of Gaussian to apply residual refinement as opposed to substitution.}
    \vspace{-3mm}
    \setlength{\tabcolsep}{0.02\linewidth}
    \resizebox{0.9\linewidth}{!}{
    \begin{tabular}{ccc|cc}
        \toprule
        Deep Supervision & Sparse Conv. & Residual Refine & mIoU & IoU \\
        \midrule
        & \checkmark & mean & 16.36  & 29.32 \\
        \checkmark &  & mean & 15.93  & 28.99  \\
        \checkmark & \checkmark & none & -  & -  \\
        \checkmark & \checkmark & all except semantics & 16.24  & 29.30 \\
        \checkmark & \checkmark & mean & \textbf{16.41}  & \textbf{29.37}  \\
        \bottomrule
    \end{tabular}}
    \vspace{-7mm}
    \label{tab:components}
\end{table*}

\textbf{Analysis of components of GaussianFormer.} 
In Table~\ref{tab:components}, we provide comprehensive analysis on the components of GaussianFormer to validate their effectiveness.
We conduct these experiments on nuScenes and set the number of 3D Gaussians to 51200.
The strategy for refinement of Gaussian properties has a notable influence on the performance.
The experiment for the all substitution strategy (denoted by none) collapses where we directly replace the old properties with new ones, and thus we do not report the result.
This is because the positions of Gaussians are sensitive to noise which quickly converge to a trivial solution without regularization for coherence during refinement.
Rectifying all properties except semantics in a residual style (denoted by all except semantics) is also harmful because the sigmoid activation used for covariance is prone to vanishing gradient.
Moreover, the 3D sparse convolution in the self-encoding module is crucial for the performance because it is responsible for the interactions among 3D Gaussians.
On the other hand, the deep supervision strategy also contributes to the overall performance by ensuring that every intermediate refinement step benefits the 3D perception.

\textbf{Effect of the number of Gaussians.}
Table~\ref{tab:number of gaussians} presents the ablation study on the number of Gaussians, where we analyze its influence on efficiency and performance.
The mIoU increases linearly when the number of 3D Gaussians is greater than 38400, which is due to the enhanced ability to represent finer details with more Gaussians.
The latency and memory consumption also correlate linearly with the number of Gaussians, offering flexibility for deployment.

\begin{figure*}[t]
\centering
\includegraphics[width=\textwidth]{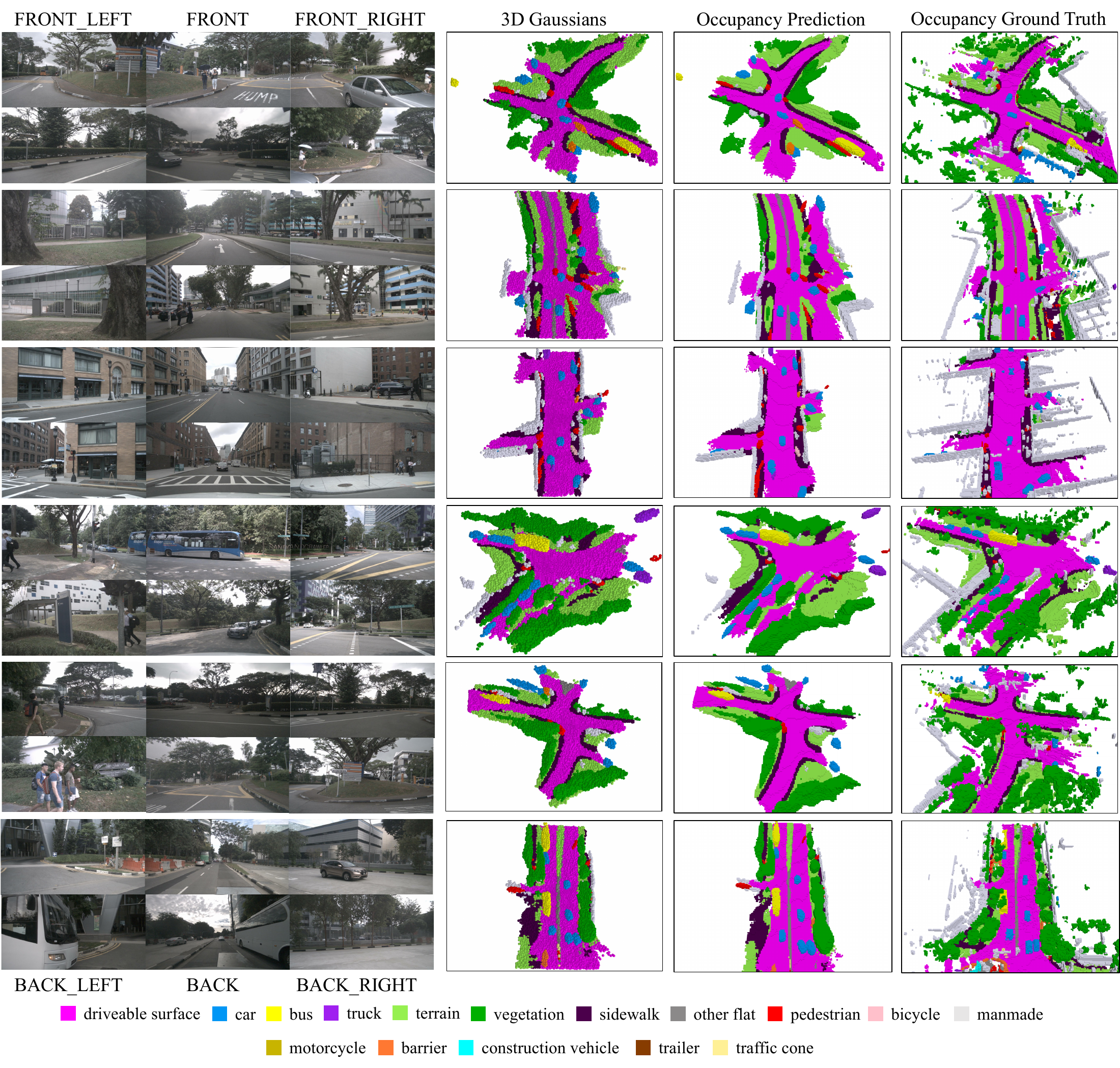}
\vspace{-7mm}
\caption{\textbf{Visualization results for 3D semantic occupancy prediction on nuScenes.}
We visualize the 3D Gaussians by treating them as ellipsoids centered at the Gaussian means with semi-axes determined by the Gaussian covariance matrices.
Our GussianFormer not only achieves reasonable allocation of resources, but also captures the fine details of object shapes.
}
\label{fig:vis}
\vspace{-7mm}
\end{figure*}

\textbf{Visualization results.}
We provide qualitative visualization results in Fig.~\ref{fig:vis}.
Our GaussianFormer can generate a holistic and realistic perception of the scene.
Specifically, the 3D Gaussians adjust their covariance matrices to capture the fine details of object shapes such as the flat-shaped Gaussians at the surface of the road and walls (e.g. in the third row).
Additionally, the density is higher in regions with vehicles and pedestrians compared with the road surface (e.g. in the first row), which proves that the 3D Gaussians cluster around foreground objects with iterative refinement for a reasonable allocation of resources.
Furthermore, our GaussianFormer even successfully predicts objects that are not in the ground truth and barely visible in the images, such as the truck in the front left input image and the upper right corner of the 3D visualizations in the fourth row.

\begin{table*}[t]
    \centering
    \caption{
    \textbf{Ablation on the number of Gaussians.}
    The latency and memory are tested on an NVIDIA 4090 GPU with batch size one during inference.
    The performance improves consistently with more Gaussians while taking up more time and memory.
    }
    \vspace{-3mm}
    \setlength{\tabcolsep}{0.032\linewidth}
    \resizebox{0.9\linewidth}{!}{
    \begin{tabular}{c|cc|cc}
        \toprule
        Number of Gaussians & Latency & Memory & mIoU & IoU \\
        \midrule
        25600  & \textbf{227} ms & \textbf{4850} M  & 16.00  & 28.72 \\
        38400  & 249 ms & 4856 M  & 16.04  & 28.72 \\
        51200  & 259 ms & 4866 M  & 16.41  & 29.37 \\
        91200  & 293 ms & 5380 M  & 18.31  & 27.48  \\
        144000 & 372 ms & 6229 M  & \textbf{19.10}  & \textbf{29.83}  \\
        \bottomrule
    \end{tabular}}
    \vspace{-7mm}
    \label{tab:number of gaussians}
\end{table*}

%% file: sections/5_conclusions.tex
\section{Conclusion and Discussions}
In this paper, we have proposed an efficient object-centric 3D Gaussian representation for 3D semantic occupancy prediction to better exploit the sparsity of occupancy and the diversity of object scales.
We describe the driving scenes with sparse 3D Gaussians each of which is characterized by its position, covariance and semantics and represents a flexible region of interest.
Based on the 3D Gaussian representation, we have designed the GaussianFormer to effectively learn 3D Gaussians from input images through attention mechanism and iterative refinement.
To efficiently generate voxelized occupancy predictions from 3D Gaussians, we have proposed an efficient Gaussian-to-voxel splatting method, which only aggregates the neighboring Gaussians for each voxel.
GaussianFormer has achieved comparable performance with state-of-the-art methods on the nuScenes and KITTI-360 datasets, and significantly reduced the memory consumption by more than 75\%.
Our ablation study has shown that the performance of GaussianFormer scales well to the number of Gaussians.
Additionally, the visualization has proved the abilities of 3D Gaussians to capture the details of object shapes and to reasonably allocate computation and storage resources.

\textbf{Limitations.}
The performance of GaussianFormer is still inferior to state-of-the-art methods despite the much lower memory consumption. 
This might result from the inaccuracy of 3D semantic Gaussian representation or simply the wrong choice of hyperparameter since we did not perform much hyperparameter tuning.
GaussianFormer also requires a large number of Gaussians to achieve satisfactory performance.
This is perhaps because the current 3D semantic Gaussians include empty as one category and thus still can be redundant.
It is interesting to only model solid objects to further improve performance and speed.

%% file: sections/6_appendix.tex
\appendix

\renewcommand\twocolumn[1][]{#1}%

\vspace{2mm}
\begin{center}
    \centering
    \includegraphics[width=\linewidth]{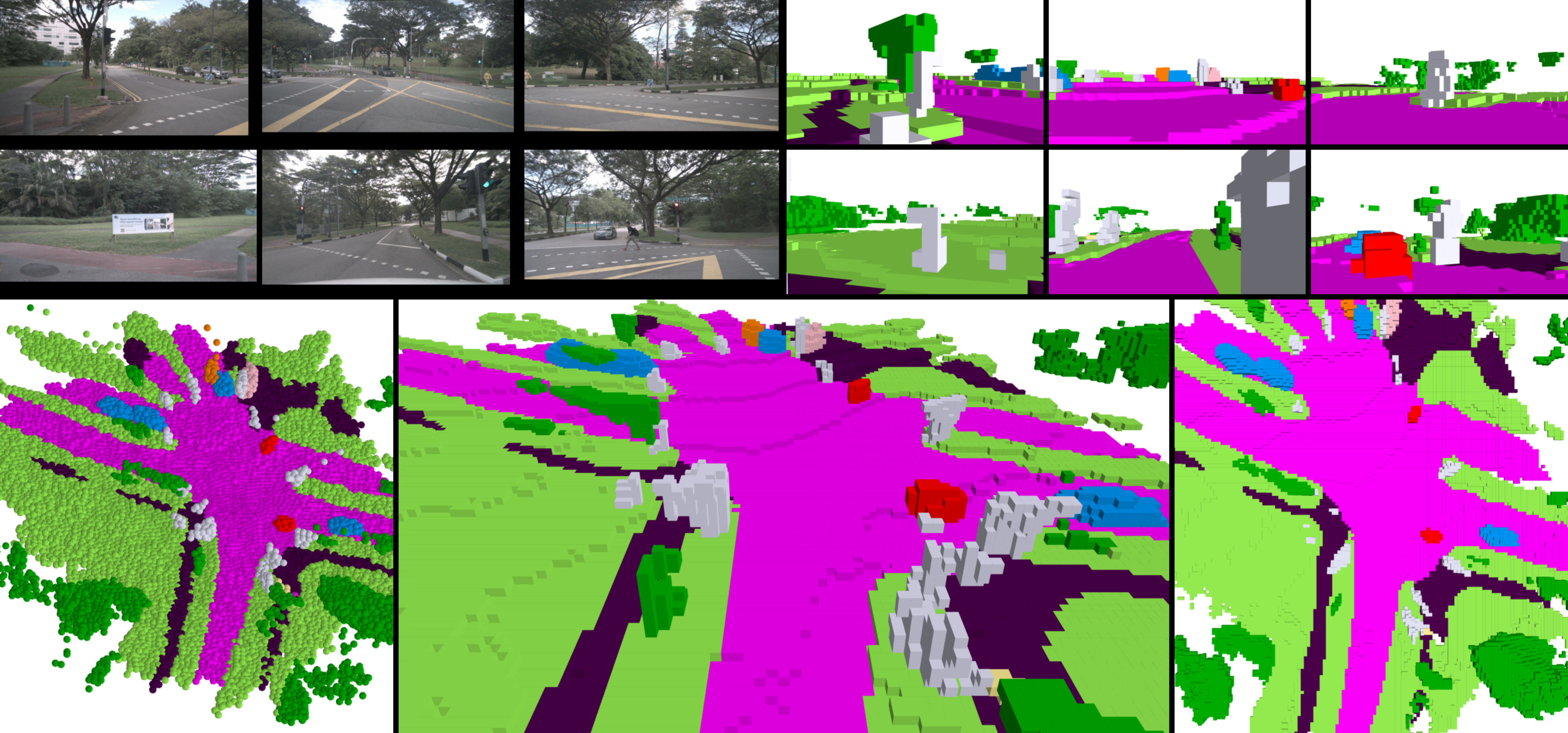}
    \vspace{-7mm}
    \captionof{figure}{
    \textbf{Visualizations of the proposed GaussianFormer method for 3D semantic occupancy prediction on the nuScenes~\cite{caesar2020nuscenes} validation set.}
    We visualize the six input surrounding images and the corresponding predicted semantic occupancy in the upper part.
    The lower row shows the predicted 3D Gaussians (left), the predicted semantic occupancy in the global view (middle) and the bird's eye view (right). 
    }
\label{supp_teaser}
    \vspace{-4mm}
\end{center}

\section{Video Demonstration}
Fig.~\ref{supp_teaser} shows a sampled image from the video demo\footnote{\url{https://wzzheng.net/GaussianFormer}} for 3D semantic occupancy prediction on the nuScenes~\cite{caesar2020nuscenes} validation set.
GaussianFormer successfully predicts both realistic and holistic semantic occupancy results with only surround images as input.
The similarity between the 3D Gaussians and the predicted occupancy suggests the expressiveness of the efficient 3D Gaussian representation.

\section{Additional Visualizations}
\begin{figure*}[t]
\centering
\includegraphics[width=\textwidth]{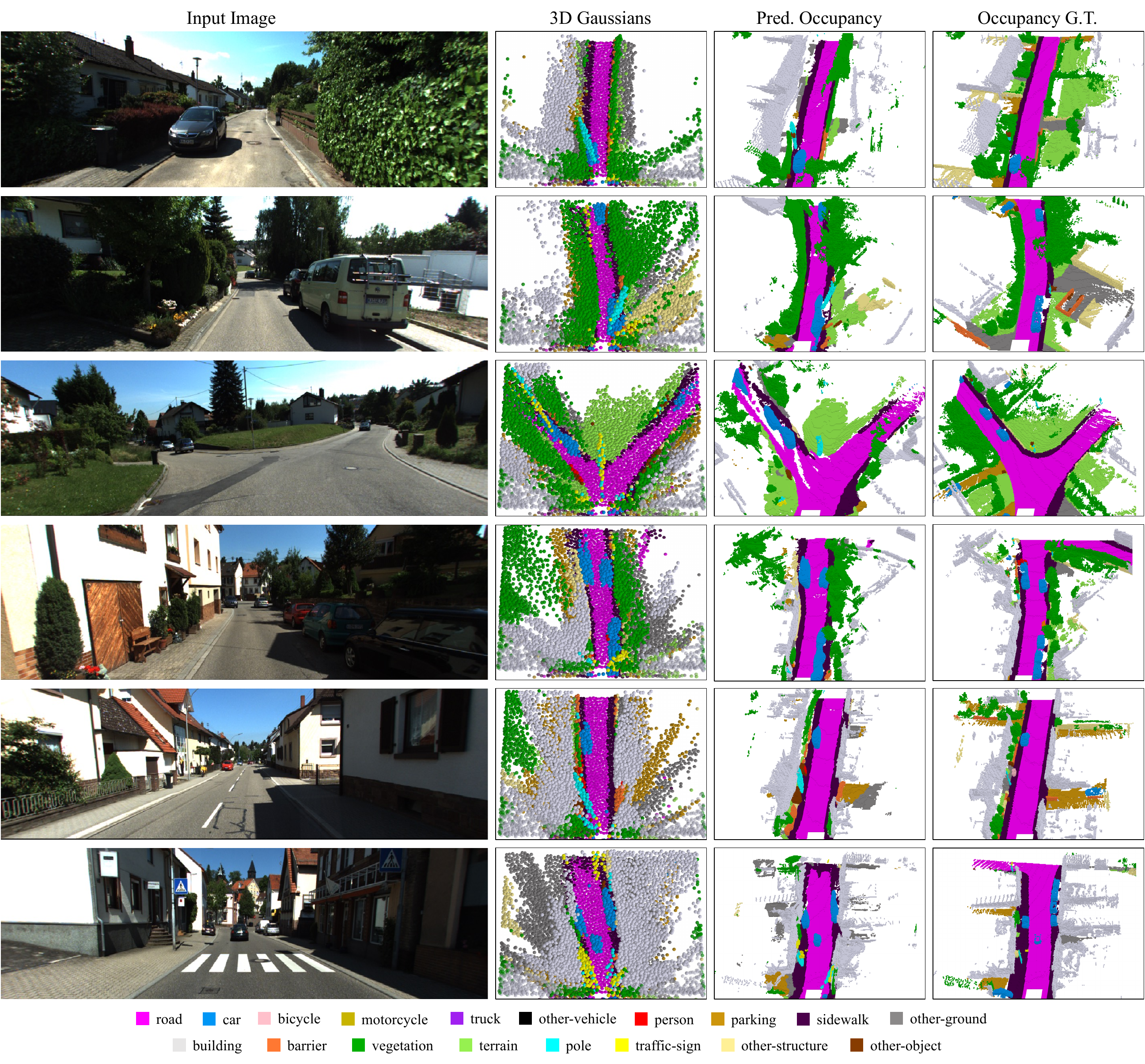}
\vspace{-7mm}
\caption{
\textbf{Visualizations of the proposed GaussianFormer method for 3D semantic occupancy prediction on the KITTI-360~\cite{Liao2022kitti360} validation set.}
GaussianFormer is able to capture both the overall structures and intricate details of the driving scenes in a monocular setting.
}
\label{app fig: kitti occ}
\vspace{-5mm}
\end{figure*}

\begin{figure*}[t]
\centering
\includegraphics[width=\textwidth]{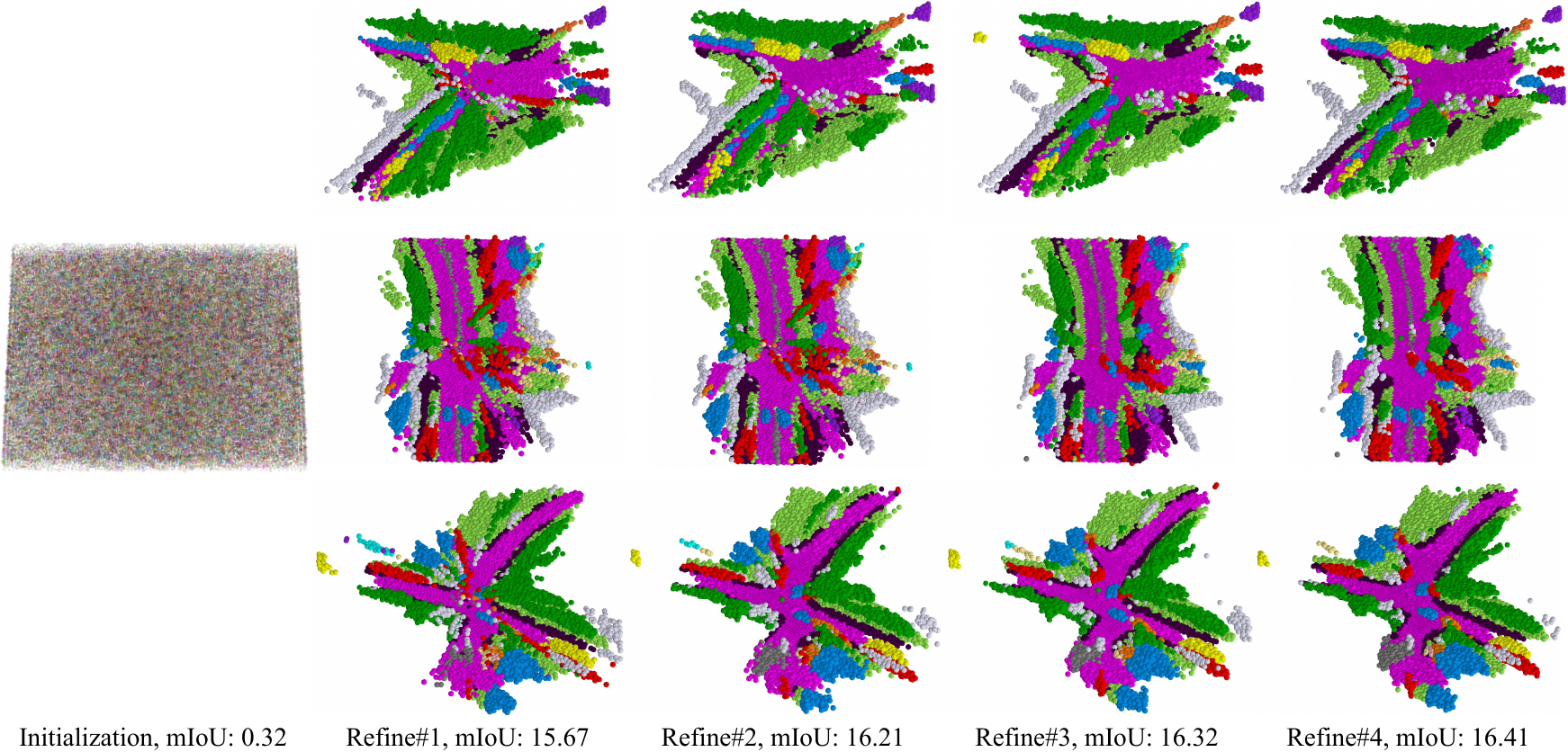}
\vspace{-6mm}
\caption{
\textbf{Visualization of the outputs of the refinement layers and the corresponding mIoU on nuScenes.}
}
\label{app fig: refine}
\vspace{-3mm}
\end{figure*}

\begin{figure*}[t!]
\centering
\includegraphics[width=\textwidth]{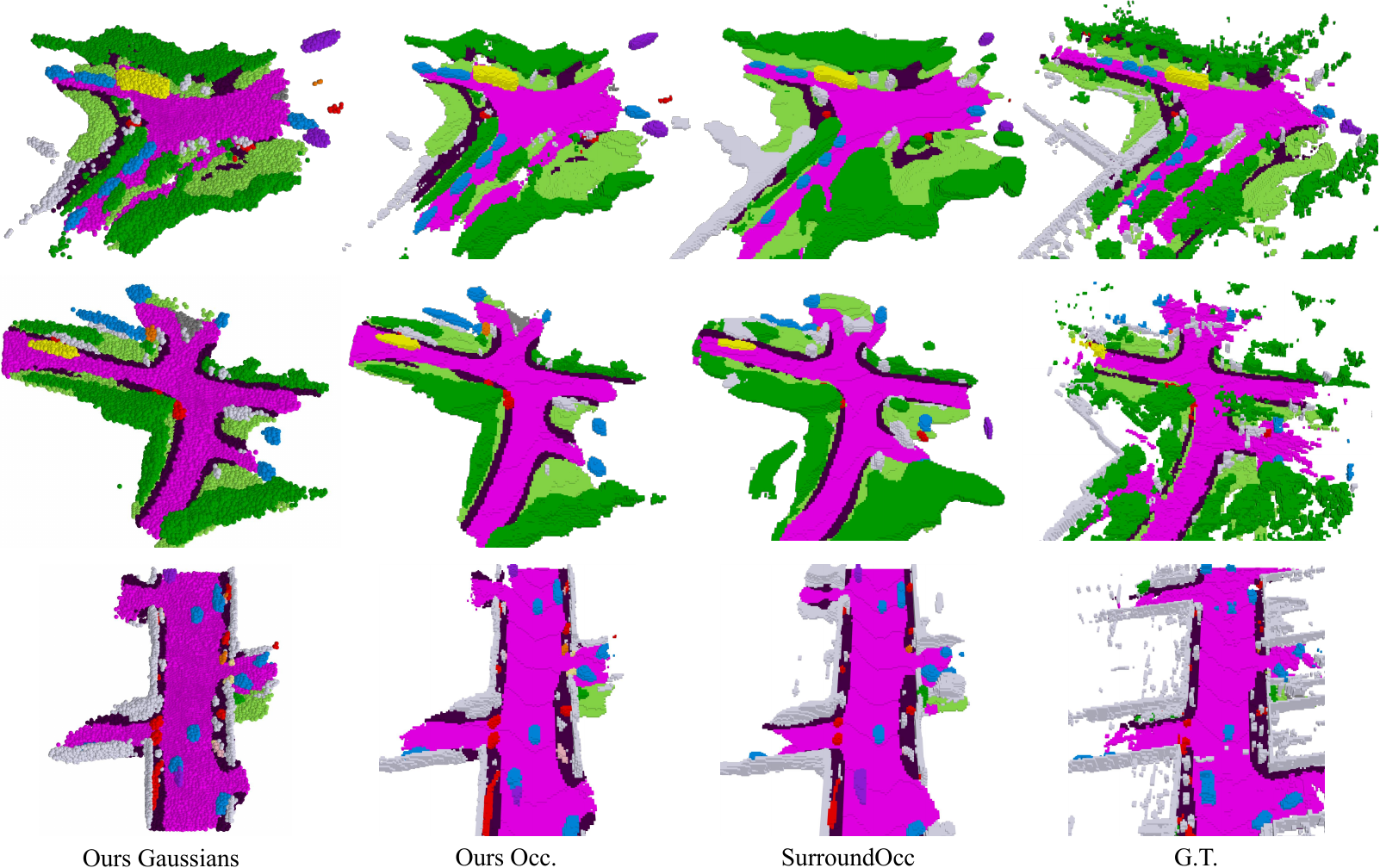}
\vspace{-6mm}
\caption{
\textbf{Qualitative comparison between our method and SurroundOcc on nuScenes.}
}
\label{app fig: compare surroundocc}
\vspace{-3mm}
\end{figure*}

In Fig.~\ref{app fig: kitti occ}, we provide the visualizations of GaussianFormer for 3D semantic occupancy prediction on the KITTI-360~\cite{Liao2022kitti360} validation set.
Similar to the visualizations for nuScenes~\cite{caesar2020nuscenes} in Fig.~\ref{fig:vis}, we observe that the density of the 3D Gaussians is higher with the presence of vehicles (e.g. the 4th row) which demonstrates the object-centric nature of the 3D Gaussian representation and further benefits resource allocation.
In addition to overall structure, GaussianFormer also captures intricate details such as poles in the scenes (e.g. the 1st row).
The discrepancy of the density and scale of the 3D Gaussians in Fig.~\ref{fig:vis} and \ref{app fig: kitti occ} is because we set the number of Gaussians to 144000 / 38400 and the max scale of Gaussians to 0.3m / 0.5m for nuScenes and KITTI-360, respectively.
We also visualize the output 3D semantic Gaussians of each refinement layer in Fig.~\ref{app fig: refine}.
In Fig.~\ref{app fig: compare surroundocc}, we provide a qualitative comparison between our GaussianFormer and SurroundOcc~\cite{wei2023surroundocc}.

\section{Additional Ablation Study}
We conduct additional ablation study on the number of Gaussians, photometric supervision, splitting \& pruning strategy, and initialization schemes in Table~\ref{tab: rebuttal exp}.
We observe consistent improvement as the number of Gaussians increases.

We also experiment with an additional photometric loss to reconstruct the input image in Table~\ref{tab: rebuttal exp}, which does not demonstrates a significant improvement.
This is because using photometric loss on nuScenes where the surrounding cameras share little overlap view cannot provide further structural information.
In addition, we experiment with the splitting and pruning strategy prevalent in the offline 3D-GS literature and observe that it improves performance compared with the baseline.

We employ learnable properties as initialization for our online model to learn a prior from different scenes in the main paper.
We further experimented with different initialization strategies including uniform distribution, predicted pseudo point cloud, and ground-truth point cloud.
We see that initialization is important to the performance and depth information is especially crucial.

\begin{table}[t]
    \centering
    \caption{
    \textbf{Analysis on different design choices.}
    }
    \vspace{-3mm}
    \setlength{\tabcolsep}{0.01\linewidth}
    \resizebox{0.9\linewidth}{!}{
    \begin{tabular}{cccc|cc}
        \toprule
        Gaussian Number & Photometric & Split \& Prune   & Initialization & mIoU & IoU \\
        \midrule
        144000 &   \ding{53}       &   \ding{53}  & learnable & {19.10} & {29.83}  \\
        192000 &       \ding{53}       &  \ding{53}   & learnable & 19.65   & 30.37    \\
        256000 &         \ding{53}     &   \ding{53}  & learnable & \textbf{19.76}   & \textbf{30.51}    \\
        \midrule
        51200  &      \ding{53}        &  \ding{53}   & learnable & {16.41} & 29.37    \\
        51200  & \color{red}$\checkmark$ &   \ding{53}  & learnable & 16.24   & \textbf{29.43} \\
        51200  &  \ding{53}  & \color{red}$\checkmark$  & learnable &  \textbf{16.50}  &  29.41  \\
        \midrule
        51200  &  \ding{53}  &       \ding{53}        & uniform   & 16.27   & 29.23   \\
        51200  &  \ding{53}  &      \ding{53}         & pseudo points & 18.99 & 28.84  \\
        51200  &   \ding{53} &        \ding{53}       & G.T. points & \textbf{26.78} & \textbf{41.81}   \\
        \bottomrule
    \end{tabular}
    }
    \vspace{-7mm}
    \label{tab: rebuttal exp}
\end{table}